\documentclass{article}

\usepackage[final]{amlc_guide}

\usepackage[utf8]{inputenc}
\usepackage[T1]{fontenc}
\usepackage{hyperref}
\usepackage{url}
\usepackage{booktabs}
\usepackage{makecell}
\usepackage{amsfonts}
\usepackage{amsmath}
\usepackage{nicefrac}
\usepackage{microtype}
\usepackage{xcolor}
\usepackage{graphicx}
\usepackage{algorithm}
\usepackage{algorithmic}
\usepackage{pgfplots}
\usepackage{tikz}
\usepackage{enumitem} 
\pgfplotsset{compat=1.18}
\usetikzlibrary{shapes,arrows,positioning,fit}
\title{K-OPSD: Verifiable On-Policy Self-Distillation for Post-Training Vision-Language Models on AEC Drawings}

\author{%
  Yunfei Bai \\
  Amazon Web Services \\
  \And
  Enrico Chionna \\
  Amazon Web Services \\
  \And
  Akash Amol \\
  Amazon Web Services \\
  \And
  Kawaljit Singh KC \\
  Amazon Web Services \\
  \And
  Joern Tinnemeyer \\
  Amazon Web Services \\
}

\begin{document}

\maketitle

\begin{abstract}

Interpreting architecture, engineering, and construction (AEC) drawings is hard for general Multimodal Large Language Models (MLLMs) and vision-language models (VLMs). We introduce K-OPSD, a VLM post-training methodology for improving AEC drawing understanding. Building on On-Policy Self-Distillation (OPSD) with verifiable supervision, we construct a teacher from the model’s own best-of-N generations, certified by a process-level verifier, and rescue failed prompts by resampling under a hint that exposes the verified answer. We then perform an on-policy model update by training on verified completions with a cross-entropy inner-loss, outperforming the bounded token-wise generalized Jensen–Shannon divergence (JSD) used by on-policy distillation. Using K-OPSD, we fine-tune Qwen3-VL models on the AECV-Bench dataset. The resulting models attain the top average judge score (0.819) and combined accuracy (0.738), achieving competitive results against open-source baseline models. The recipe transfers to the out-of-domain ArchCAD dataset, where the 8B model gains most. We present the verifier suite and the continual learning and self-improving pipeline, our results provide preliminary evidence that verifier-guided self-distillation is a promising route toward more reliable machine reading of architecture drawings.
  
\end{abstract}

\section{Introduction}

Interpreting architecture drawings, such as 2D floor plans, sections, and construction details, together with their 3D Building Information Modeling (BIM) counterparts, is a foundational capability for the Architecture, Engineering, and Construction (AEC) domain, yet it remains hard for conventional computer-vision methods and for general Multimodal Large Language Models (MLLMs) and Vision-Language Models (VLMs) alike: drawings encode dense symbolic conventions, precise geometry, and implicit engineering semantics that these models routinely misread. Recent AEC VLMs for component-to-BIM synthesis and floor-plan vectorization, and multimodal reasoning models for structural drawings and damage assessment, show the promise of the approach, but rely on supervised training alone and none exploits the fact that correctness here is largely machine-verifiable. We observe that understanding architecture drawings is uniquely suited to verifiable supervision, because a model’s extracted geometry and semantics can be checked deterministically against ground truth—dimensional facts, symbol and entity libraries, BIM schema and geometric validity, and building-code rules. Yet supervised fine-tuning (SFT) maximizes likelihood on gold traces off-policy, leaving an exposure-bias gap between teacher-forced training and inference, and it cannot distinguish a correct derivation from a merely plausible one. Reinforcement learning with verifiable rewards (RLVR) also exploits checkable correctness, but does so through a scalar trajectory-level reward that supplies no per-token target. On-policy self-distillation (OPSD) instead converts verified correctness into dense token-level supervision, and it is this regime that we study and improve. The model learns from its own on-policy generations under a token-level objective, keeping check-certified samples and distilling them back into the weights (STaR [21], rejection-sampling fine-tuning [22], the ReST family [23, 24], generalized knowledge distillation (GKD) [25]). However, OPSD’s bounded token-wise generalized Jensen–Shannon inner loss saturates in gradient precisely on high-confidence wrong tokens, and it depends on a teacher of uneven quality: privileged-information teacher prompts bias the student toward the reference and produce "right answer, wrong reasoning" traces whose per-token loss drops while validation accuracy stalls or degrades [26, 27].

To address these challenges, we introduce K-OPSD, an adaptive OPSD recipe that threads verifiable supervision through the entire adaptation stack. K-OPSD builds a self-teacher from the model’s own verified best-of-N generations: for each prompt, the model produces multiple candidate responses scored by a deterministic verifier, then resamples the low-quality prompts under hint-conditioned guidance for further verification. The resulting verified corpus forms the teacher toward which the student is distilled. The self-teacher is first placed on an SFT-trained base model to establish a competent initialization, followed by an on-policy weight update that follows the gradient of the cross-entropy inner loss on the verified teacher token sequence, rather than the bounded divergence of the teacher’s probability distribution. We use this pipeline to post-train small Qwen3-VL base models for AEC drawing understanding. Throughout, the fine-tuned VLM optimizes a composite verifiable objective spanning geometric accuracy, schema conformance, symbol recognition, and regulatory consistency. Our key contributions include:

\begin{itemize}  [leftmargin=0.5cm]
  \item We introduce K-OPSD, a verifiable-guided on-policy self-distillation post-training loop. Without requiring an external teacher model, the pipeline continually builds its own teacher from the model’s own policy through a process-level verification gate covering perception, reasoning, and the final answer.
  \item We improve the OPSD gradient update through cross-entropy loss on the verified completions, outperforming the bounded token-wise generalized JSD. In our empirical experiments, we show that verification-gated selection, not soft distribution matching, carries the strongest signal for distillation.
  \item Using K-OPSD, we post-train small Qwen3-VL base models for the visual question answering (VQA) task of reading architecture drawings. Our empirical experiments show competitive results in accuracy against Pixtral, Kimi, Gemma-3, Llama4, and other foundation models, demonstrating that a small domain-specialized model can reach large-model quality on the VQA task.
  \item Our self-improving and continual learning pipeline offers a practical recipe that, in our experiments, shows promise as a project-agnostic paradigm for more reliable machine reading of architecture drawings.
\end{itemize}

%\vspace{-10pt} 
\section{Related work}

Our work sits at the intersection of two research threads: the emerging application of LLMs to the AEC domain, and the line of self-taught reasoning and on-policy distillation that lets a model improve without an external teacher. 

\textbf{LLM applications in the AEC domain} Design authoring and BIM generation. LLMs increasingly generate and manipulate BIM: Text2BIM [3] and a speech-to-BIM framework [8] translate natural language into authoring commands, while BIMgent [13], BIMVLM [16], and FLOORPLANVLM [17] push toward autonomy and multimodal, visual-input synthesis. All establish LLMs as capable design-authoring engines, but rely on prompting, agentic orchestration, or supervised training. None learns from verification of its own generations. Automated interpretation of building codes is a recurring target: prior work covers LLM-based regulation interpretation [7], compliance checking against BIM data [5], and automated code review [9], while surveys [19, 20] confirm reliable regulatory reasoning remains open. Notably, correctness here is often formally checkable against explicit rules, a property our work exploits but which prior compliance studies use only for post-hoc evaluation. Beyond design and compliance, LLMs have been applied to structural analysis with an emphasis on reliability and robustness [10], structural drawing generation via ReAct-style reasoning and retrieval-augmented generation (RAG) [11], multimodal structural damage identification with chain-of-thought reasoning [12], and geotechnical foundation design through router-based multi-agent architectures [14]. Design-quality assessment has likewise been explored, for example by processing design inspection reports with LLMs [18]. Across these works, domain adaptation almost never advances beyond prompting, RAG, agentic scaffolding, or, at most, supervised fine-tuning [1, 12]. 

\textbf{Domain models and benchmarks} A smaller set of works adapts model weights or measures domain competence directly. Qwen-BIM [1] fine-tunes a Qwen-based model for BIM design tasks with an accompanying dataset and benchmark, and a comparative study [2] weighs domain-specific fine-tuning against prompt-based learning for BIM information retrieval. On the evaluation side, CEQuest [4] benchmarks construction estimation, DrafterBench [6] benchmarks civil-engineering task automation, and an examination of fire-engineering technical questions [15] quantifies the gap between general LLMs and domain knowledge. Moving to the visual modality, AECV-Bench [28] benchmarks multimodal models on architectural and engineering drawing understanding, and ArchCAD-400K [29] supplies a large-scale CAD-drawing dataset and baseline for panoptic symbol spotting. Collectively, these benchmarks reveal substantial shortfalls in engineering-grade quality, yet they are consistently used as evaluation harnesses rather than as training signals.

\textbf{Self-taught reasoning and on-policy distillation} Our pipeline draws on work in which a model improves from its own generations. STaR [21] establishes the template of keeping sampled traces that reach the correct answer, and rationalize failures by conditioning on the known answer to recover a trace that is then used for training after removing the hint, while RFT [22] scales the "keep verified-correct samples" half and the ReST family [23, 24] casts it as an iterated grow-then-improve loop where verifier-filtered data beats human data. On how to distill self-generated data, GKD [25] trains on the student's own samples under a token-wise generalized Jensen-Shannon divergence, but recent analyses caution that answer-conditioned (privileged-information) teachers bias the student toward the reference [26] and that on-policy self-distillation needs rethinking for thinking models [27]. 

%\vspace{-10pt} 
\section{Methodology}

\subsection{Training pipeline}

\begin{figure}  [htbp]
    \centering
    \includegraphics[width=1.0\textwidth]{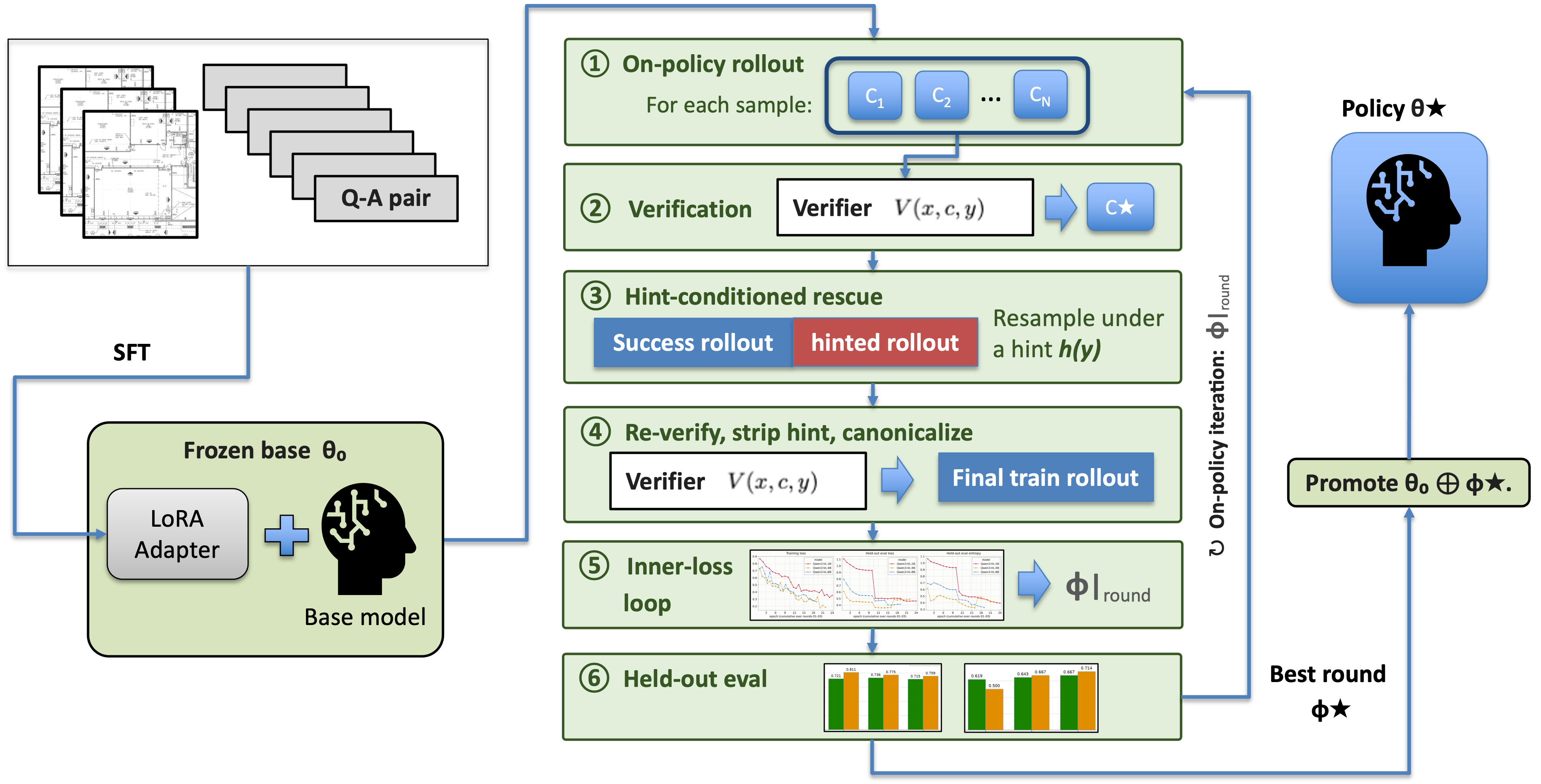}
    \caption{\textbf{K-OPSD Training Paradigm.} First, SFT establishes a format-conformant base on SME-validated, correctness-filtered ground truth, which is then frozen for all subsequent training. Each round of verifiable on-policy self-distillation samples best-of-$N$ completions from the current student, keeps only those a process-level verifier certifies, and rescues unsolved prompts by resampling under a hint that is stripped and re-verified before the trace becomes a target. Then a fresh LoRA adapter is fit on the frozen SFT-base over the verified corpus by cross-entropy inner loss and evaluated on a held-out split. After $R$ rounds the best-scoring adapter is promoted as the final model. }
    \label{fig:train_flow}
\end{figure}

We consider a dataset $\mathcal{D} = \{(x, y)\}$ of drawing-understanding instances, in which each prompt $x$ pairs a rendered architecture drawing with a question and each $y$ is its ground-truth answer (a dimensional fact, a symbol or entity label, a BIM-schema value, or a building-code judgment). The learner is a vision-language policy $\pi_\theta$ that, given $x$, produces a completion $c = (r, \hat{y}) \sim \pi_\theta(\cdot \mid x)$ pairing a reasoning trace $r$ with an extracted answer $\hat{y}$. The staged pipeline is described in Figure~\ref{fig:train_flow}.

\textbf{Step 1: On-policy rollout} The student policy is the frozen base composed with the previous adapter $\pi_r \;:=\; \pi_{\theta_0 \oplus \phi_{r-1}}$, the first round samples from the bare SFT base $(\phi_0 = \emptyset)$. For every prompt $(x, y) \in \mathcal{D}$, we draw $N$ independent completions from this policy,
\begin{equation}
\mathcal{C}(x) \;=\; \big\{\, c_i = (r_i, \hat{y}_i) \sim \pi_r(\cdot \mid x) \;:\; i = 1, \dots, N \,\big\}.
\end{equation}
The supervision assembled targets the exposure-bias gap between teacher-forced training and free-running inference directly.

\textbf{Step 2: Verification gate} Each completion is scored by the deterministic, process-level verifier $V$, and the passing subset is retained,
\begin{equation}
\mathcal{C}^{+}(x) \;=\; \big\{\, c \in \mathcal{C}(x) \;:\; V(x, c, y) = 1 \,\big\}.
\end{equation}
When $\mathcal{C}^{+}(x) \neq \emptyset$, we keep a single verified target $c^\star = \mathrm{BestOf}\big(\mathcal{C}^{+}(x)\big)$ and add $(x, c^\star)$ to the round's teacher corpus. $V$ selects among the student's own outputs and ground truth $y$ is never used directly as a label. Because $V$ checks the trace $r$ rather than only the extracted answer $\hat{y}$, it prevents superficially-correct answers from passing.

\textbf{Step 3: Hint-conditioned rescue of unsolved prompts} The prompts with $\mathcal{C}^{+}(x) = \emptyset$ are not discarded, instead we resample once under a hint $h(y)$ that exposes the gold answer and a teacher reasoning trace 
\begin{equation}
\tilde{c} \;\sim\; \pi_r\big(\cdot \mid x, \, h(y)\big)
\end{equation}
where $h(y)$ is visible in the prompt at sampling time only. Conditioning on $h(y)$ raises the probability of a correct trajectory on precisely the hard cases, recovering coverage that would otherwise cap self-training at problems already within reach.

\textbf{Step 4: Re-verification, hint stripping, and canonicalization} A rescued trace is admitted only if it independently passes the same verifier and clears a leakage guard $V(x, \tilde{c}, y) = 1 \;\;\text{and}\;\; \lnot\,\mathrm{Leaked}(\tilde{c})$, 
after which the hint is removed and the completion is canonicalized to an explicit reasoning-then-answer form,
\begin{equation}
c \;=\; \mathrm{StripHint}(\tilde{c}) \;=\; \langle\texttt{think}\rangle\, r\, \langle/\texttt{think}\rangle\,\langle\texttt{answer}\rangle\, \hat{y}\, \langle/\texttt{answer}\rangle.
\end{equation}
Completions lacking a $\langle\texttt{think}\rangle$ segment are dropped. Steps 2–4 together produce the round's verified teacher corpus,
\begin{equation}
\mathcal{T}_r \;=\; \bigcup_{(x, y) \in \mathcal{D}} \big\{ (x, c) \big\}
\end{equation}
in which every target is the student's own trace and the answer $y$ never appears in the conditioning. 

\textbf{Step 5: Inner-loop optimization of a fresh adapter} A new low-rank adapter is fitted on the frozen base over the round's corpus,
\begin{equation}
\phi_r \;=\; \arg\min_{\phi} \; \frac{1}{|\mathcal{T}_r|} \sum_{(x, c) \in \mathcal{T}_r} \mathcal{L}\big(\theta_0 \oplus \phi; \, x, c\big)
\end{equation}
with early stopping selecting the lowest-evaluation-loss checkpoint as $\phi_r$. The inner loss $\mathcal{L}$ is is cross-entropy over the verified completion.

\textbf{Step 6: Held-out evaluation against a same-round baseline} The updated policy is scored on a stratified held-out split $s_r \;=\; \mathrm{Eval}\big(\theta_0 \oplus \phi_r, \, \mathcal{D}_{\mathrm{val}}\big)$ and compared against the same-round pre-update policy $\pi_r$,
\begin{equation}
\Delta_r \;=\; s_r - \mathrm{Eval}\big(\theta_0 \oplus \phi_{r-1}, \, \mathcal{D}_{\mathrm{val}}\big)
\end{equation}
so that the measured gain isolates the round's own contribution and is not confounded by cross-round drift in the evaluation conditions.

\subsection{The process-level verifier}

The defining property of this domain is that correctness is machine-verifiable, where $V$ is a composite gold-type-routed verifier for a completion $c$. It computes per-family scores over the drawing and aggregates them into a scalar quality score $s (x, c, y) \in [0,1]$, and emits the binary gate decision $V(x, c, y) = 1[ s (x, c, y) \geq \tau]$ with an acceptance threshold $\tau$ that the user can configure. The gate is therefore hard-binary at the point of selection, while the underlying score is soft. Beyond checking whether $\hat{y}$ matches $y$, $V$ verifies that the trace $r$ is consistent with the drawing dimensionally, schematically, symbolically, and against code, which is what makes the verification "process-level" rather than answer-only. 

\subsection{The self-distillation}

Post-training seeks parameters $\theta$ that maximize the expected verified correctness $\mathbb{E}_{x \sim \mathcal{D}}\,\mathbb{E}_{c \sim \pi_\theta(\cdot \mid x)}\big[V(x, c, y)\big]$ of the free-running policy at inference, rather than merely its likelihood under a fixed reference corpus. This objective exposes the limitations of the standard on-policy distillation (OPD) adaptation routes and motivates our design. External-teacher distillation is impractical in our setting as a larger AEC-native teacher is not available. We therefore construct a teacher-free training signal that is simultaneously on-policy, drawn from the model's own current output distribution. The teacher is verification-gated so that every target is certified correct by $V$ rather than merely by proximity to a reference, then internalized throughout the iterated distillation by cross-entropy gradient optimization. 

The hint is used only during answer-conditioned generation, and then re-verification with the hint removed keeps it out of supervision. The resulting trace must be independently re-certified by the process-level verifier and clear a leakage guard that rejects any trace copying, referencing, or unsupported by the answer. What is distilled is therefore the model's own drawing-grounded derivation, so the hint does not lower the bar an accepted trace must clear. 

%\vspace{-10pt} 
\section{Empirical experiments}

\subsection{Experimental setup}

\textbf{Datasets}  The Qwen3-VL foundation models are post-trained on the AECV-Bench [28] dataset, first by SFT with 100 samples, followed by K-OPSD on 150 self-distillation samples. The trained models are evaluated on two held-out sets: the in-domain AECV-Bench test split (42 items) and, as an out-of-domain generalization test, ArchCAD dataset [29] (40 items). ArchCAD is never seen in training, so it measures transfer and generalization across drawing types and conventions.

\textbf{Model baseline} We situate the fine-tuned model against a reference group of open-source MLLM and VLM, listed in Table ~\ref{tab:benchmark}. The fine-tuned 4B and 8B entries are Qwen3-VL-4B and -8B post-trained with our K-OPSD recipe. Every model in Table ~\ref{tab:benchmark}, including ours, is invoked zero-shot with the same prompt and decoding settings, no model receives in-context examples or retrieval. The intended contrast is between domain knowledge acquired through post-training and domain knowledge available from large-scale pretraining, with inference held constant. The benchmark establishes that post-training a 4B–8B model on verified in-domain supervision reaches or exceeds the zero-shot performance of larger open-source models on this task, under an identical inference protocol. It does not compare K-OPSD with those models' own post-training, nor claim that they would not improve under comparable domain adaptation.

\textbf{Evaluation metrics} AEC drawing answers are expressed in free-form natural language, so a purely deterministic evaluation can penalize answers that are correct but phrased or formatted differently, and does not evaluate reasoning. Therefore, we use two accuracy metrics to measure the training performance: Combined Accuracy and Average Judge Score. The two metrics deliberately measure different constructs. Combined Accuracy is a deterministic offline metric providing an objective and reproducible measure of the final answer string alone. The judge score is an LLM-judge metric capturing the natural-language answer quality and reasoning that deterministic matching misses. It provides holistic measurement including answer alignment, response format, and whether the answer follows from the reasoning. Because only the first component is shared, the two are not expected to agree in level, and we therefore compare models by ranking and by within-metric deltas rather than across-metric magnitudes.
 
$\text{Combined  Accuracy} = \big\{0.5 \times \text{lexical token F1} + 0.5 \times \text{Semantic Similarity}\big\}$, where the semantic similarity is calculated through the all-MiniLM-L6-v2 embedding model.

$\text{Average Judge Score} = \big\{ 0.75 \times \text{answer accuracy} + 0.1 \times \text{answer format} + 0.15 \times \text{answer reasoning} \big\}$

 To reduce single-model and single-family bias in the judge, we use two independent judges drawn from different model families, glm-5 and Nova 2 Lite, to provide the rubric-based judge scores, and average their per-question scores. Each judge rates the response as a rubric-weighted sum over answer accuracy, so that a correct final answer, a sound derivation, and a well-formed response each contribute to the score. The LLM-judge is not asked to assess drawing correctness from its own domain knowledge, it rather receives the SME-validated ground-truth answer and rates alignment with it. This reference-grounded component carries 0.75 of the rubric weight, the judge's role is therefore restricted to a linguistic judgment, recognizing semantic equivalence between a free-form response and a human-validated answer, rather than an engineering one. The remaining weight covers response format (0.1), which is mechanically checkable, and whether the stated answer follows from the model's reasoning (0.15).

\textbf{Implementation details} The policy training for all models and methods use LoRA adapter on the frozen SFT initialized model, loaded in 4-bit QLoRA with bf16. The adapter geometry is held identical across the experiments: rank $r = 16$, $\alpha = 32$, dropout $0.05$, no bias. The outer self-distillation loop runs multiple rounds, with a stratified 15\% slice held out for round selection. On verified supervision, each round samples $N = 8$ completions per prompt (best-of-8) under stochastic decoding (temperature $1.0$, top-p $0.95$, up to 512 new tokens). The deterministic gold-type-routed selector applies the verifier $V$, keeping a completion when the composite verifier score satisfies $s(x, c, y) \geq \tau$ with $\tau=0.65$, and drops degenerate (not specified or empty) targets outright. Prompts for which best-of-8 fails are resampled once through the hint-conditioned rescue path (best-of-4). The inner loop of OPSD, VS-OPSD and K-OPSD run for up to 10 epochs with early stopping callback (patience 3 on eval loss). A stratified 20\% of each round's distill rows is held out for per-epoch evaluation. Rollout, hint-conditioned rescue, and the inner-loop on-policy weights update are each sharded across 4 GPUs with data parallelization, while candidate selection, gating, and metric aggregation stay single-process.

\begin{table} 
    \centering
\caption{\textbf{Fine-tuned VLMs versus frontier and open-source MLLM/VLMs}, on the in-domain AECV-Bench and out-of-domain ArchCAD sets. FT-Qwen3-VL-4B and -8B are models post-trained with our K-OPSD recipe. The FT-Qwen3VL-4B leads AECV-Bench accuracy and FT-Qwen3VL-8B leads ArchCAD accuracy, at a fraction of the baselines' parameters.}
\label{tab:benchmark}
    \begin{tabular}{llcccc}\toprule
         &  &  \multicolumn{2}{c}{AECV-Bench}&  \multicolumn{2}{c}{ArchCAD}\\
         %\cline{3-4} \cline{5-6}
         \cmidrule(lr){3-4} \cmidrule(lr){5-6} 
         &  &  \makecell{Avg Judge \\ Score}&  \makecell{Combined \\ Accuracy}&  \makecell{Avg Judge \\ Score}& \makecell{Combined \\ Accuracy}\\\midrule
 Open-source model & Qwen3-VL-235B& 0.698& 0.667& 0.538&0.268\\
 & Pixtral-Large-2502& 0.687& 0.643& 0.514&0.293\\
 & Kimi-K2.5& 0.765& 0.714& 0.619&\textbf{0.415}\\
 & Gemma-3-27B& 0.731& 0.714& 0.543&0.268\\
 & Llama4-Maverick-17B& 0.795& 0.738& 0.478&0.171\\
 & Nemotron-Nano-12B& 0.552& 0.310& 0.239&0.317\\ 
 & InternVL3.5-8B& 0.686& 0.643& 0.548&0.390\\
 & MiniCPM-V-4.6& 0.374& 0.286& 0.276&0.317\\\midrule
 \textbf{\makecell{FT-VLM (Ours)}}& \textbf{FT-Qwen3-VL-4B}& \textbf{0.819}& \textbf{0.738}& 0.660&0.390\\ 
 & \textbf{FT-Qwen3-VL-8B}& 0.811& 0.690& \textbf{0.693}&0.366\\ \bottomrule
    \end{tabular}
\end{table}

\begin{figure} [htbp]
    \centering
    \includegraphics[width=1.0\textwidth]{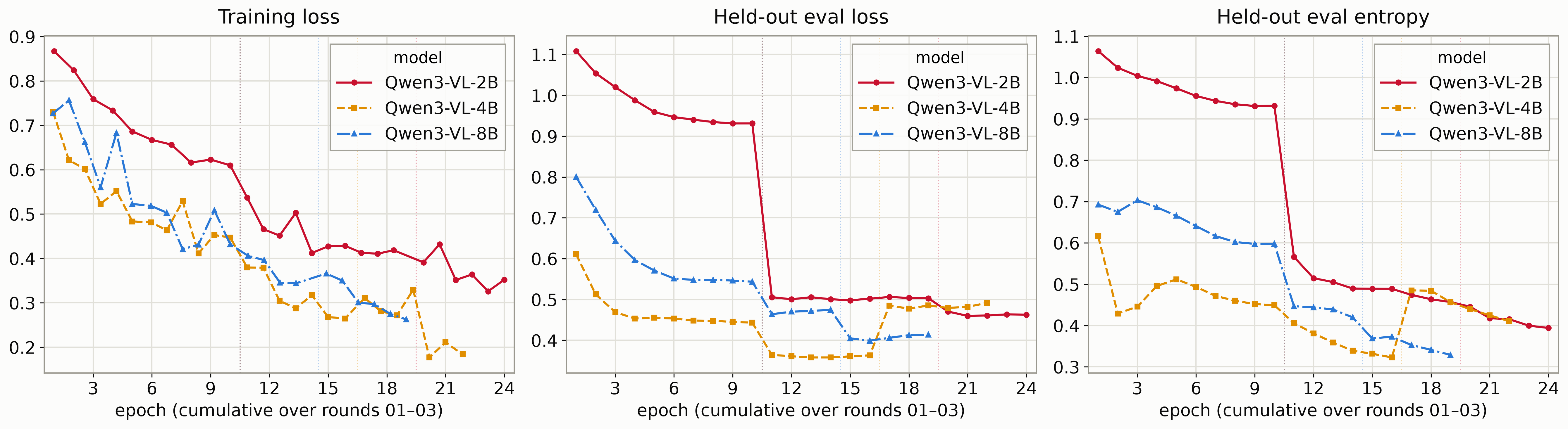}
    \caption{\textbf{K-OPSD training dynamics across rounds.} Training loss (left), held-out eval loss (middle), and held-out generation entropy (right) over training rounds for Qwen3-VL-2B, 4B, and 8B. All three quantities decline for every scale, with step drops at round boundaries as each round fits a fresh adapter on the frozen SFT base. Held-out loss and entropy fall together indicating the verified self-teacher steadily lowers error while the policy grows more confident. The entropy trace is monitored for distribution collapse across rounds.}
    \label{fig:train_process}
\end{figure}

\vspace{-10pt} 
\subsection{Fine-tuned model accuracy benchmarking}

Figure~\ref{fig:train_process} shows the smooth convergence across training rounds and epochs. As shown in Table~\ref{tab:benchmark}, On the in-domain AECV-Bench, FT-Qwen3-VL-4B attains both the highest Average Judge Score in the table (0.819) and top-tier Combined Accuracy (0.738, tying the strongest baselines Llama4-Maverick-17B and Gemma-3-27B); the two metrics agree here, which makes the in-domain result robust. On the out-of-domain ArchCAD set, FT-Qwen3-VL-8B leads all models on judge score (0.693, above the best baseline Kimi-K2.5) and improves Combined Accuracy over the base model (though it does not dominate every baseline on accuracy), evidence that the post-training transfers beyond the training domain rather than overfitting it. The two signals do not move in lockstep everywhere, at 8B in-domain the judge score is high (0.811) while Combined Accuracy (0.690) trails the top baselines. This divergence is expected given what each metric measures. The judge additionally credits reasoning and formatting quality, while Combined Accuracy rewards lexical and embedding overlap and can under-credit a correct answer phrased or structured differently. Therefore we report both metrics side by side, and read a result as strongest where the two concur. 

%\vspace{-10pt} 
\subsection{Ablation study}

For the controlled ablation, we fine-tune Qwen3-VL at 2B, 4B, and 8B under a ladder of training methods that isolate each ingredient. The ladder separates three axes: \verb|OPD → OPSD| swaps a strong external teacher (using Claude Fable-5) for the model's own on-policy self-teacher, \verb|OPSD → VS-OPSD| adds the best-of-\textit{N} process-level verification gate, and \verb|VS-OPSD → K-OPSD| changes the inner loss, giving a clean controlled ablation of generalized JSD against cross-entropy on an identical verified corpus. The method ladder is evaluated on two evaluation sets: AECV-Bench (in-domain) and ArchCAD (out-of-domain), in Table~\ref{tab:ablation}.

\begin{table}  [htbp]
    \centering
\caption{\textbf{Method-ladder ablation} on AECV-Bench and ArchCAD for Qwen3-VL at 2B, 4B, and 8B under each training stage: base, OPD, OPSD, VS-OPSD, K-OPSD. The ladder isolates three axes: external teacher vs. self-teacher, the verification gate, and the inner loss, showing that K-OPSD (cross-entropy on the verified corpus) is competitive across the methods. }
\label{tab:ablation}
    \begin{tabular}{lcccccc}\toprule
         &\multicolumn{2}{c}{Qwen3-VL-2B}&\multicolumn{2}{c}{Qwen3-VL-4B}&  \multicolumn{2}{c}{Qwen3-VL-8B}\\
         \cmidrule(lr){2-3} \cmidrule(lr){4-5} \cmidrule(lr){6-7}
 &\makecell{Avg Judge \\ Score}& \makecell{Combined \\ Accuracy}& \makecell{Avg Judge \\ Score}& \makecell{Combined \\ Accuracy}& \makecell{Avg Judge \\ Score}&\makecell{Combined \\ Accuracy}\\\midrule
         AECV-Bench&&  &  &  &  &\\
         \cmidrule(lr){1-1}
 Base model& 0.417& 0.619& 0.771& 0.738& 0.738&0.643\\
         OPD&0.665&  0.595&  0.780&  0.714&  0.792&\textbf{0.738}\\
         OPSD&0.588&  0.476&  0.713&  0.690&  0.760&0.690\\
         VS-OPSD&0.676&  0.667&  0.730&  0.667&  0.736&0.690\\
         \textbf{K-OPSD  (Ours)}&\textbf{0.699}&  \textbf{0.690}&  \textbf{0.819}&  \textbf{0.738}&  \textbf{0.811}&0.690\\ \midrule
 ArchCAD& & & & & &\\
 \cmidrule(lr){1-1}
 Base model& 0.327& 0.317& 0.621& 0.366& 0.521&0.220\\
 OPD& 0.543& 0.268& 0.660& \textbf{0.463}& 0.675&0.366\\
 OPSD& 0.497& 0.268& 0.612& 0.366& 0.580&0.317\\
 VS-OPSD& 0.497& 0.293& \textbf{0.666}& 0.439& 0.599&0.317\\
 \textbf{K-OPSD  (Ours)}& \textbf{0.589}& \textbf{0.341}& 0.660& 0.390& \textbf{0.693}&\textbf{0.366}\\ \bottomrule
    \end{tabular}
    
\end{table}

\textbf{Self-teacher ablation: model self-generation can match or exceed strong external teacher}

The \verb|OPD → OPSD| step shows that un-gated self-distillation is not enough on its own: OPSD sits below the external-teacher baseline OPD on both metrics and both sets at every scale (e.g. AECV-4B 0.713/0.690 vs 0.780/0.714, ArchCAD-8B 0.599/0.317 vs 0.675/0.366). Adding the verification gate and the cross-entropy inner loss reverses this: K-OPSD (self-taught) overtakes OPD (taught by Claude Fable 5) on judge at all three scales on AECV (0.699/0.819/0.811 vs 0.665/0.780/0.792) and matches or beats it on ArchCAD judge (2B 0.589 vs 0.543, 4B tied at 0.660, 8B 0.693 vs 0.675). Notably, the external teacher retains an accuracy edge on Combined Accuracy (AECV-8B, 0.738 vs 0.690, and ArchCAD-4B, 0.463 vs 0.390), while in most cases the verification-gated self-distillation recovers, and usually exceeds, the benefit of a strong external teacher without requiring one. These results suggest that, in this setting, a strong external teacher is not necessary, which motivates our confinement of conditioning to the re-verified rescue path.

\textbf{Verification gate ablation: verified completion helps across scale and set}   

\begin{figure}  [htbp]
    \centering
    \includegraphics[width=1.0\textwidth]{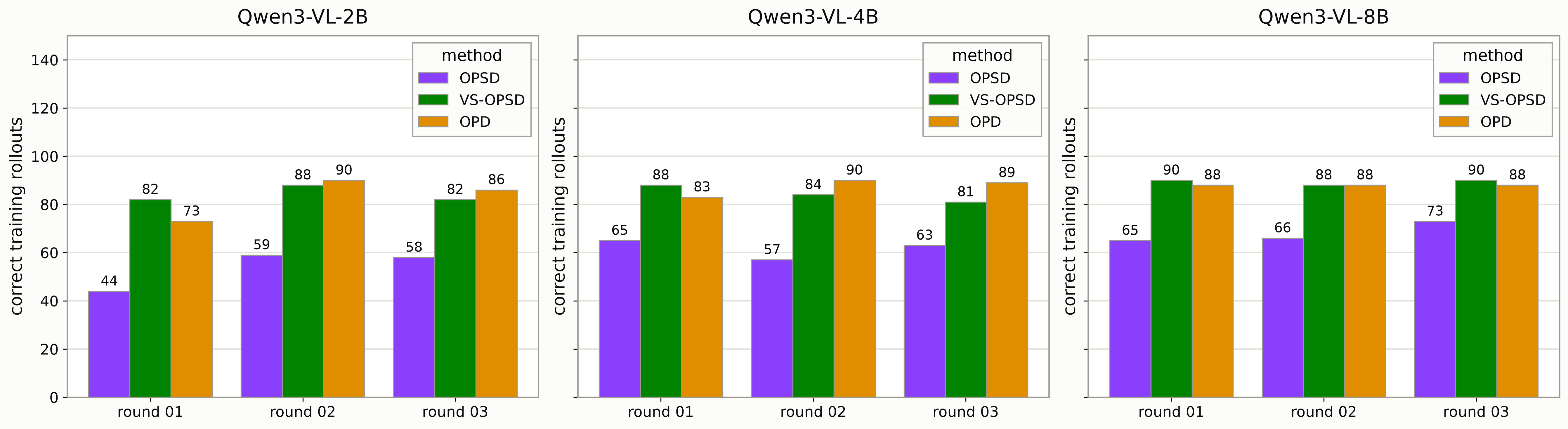}
    \caption{\textbf{Verified training-target yield by OPSD vs VS-OPSD vs OPD} across three rounds for Qwen3-VL-2B, 4B, and 8B. At every scale and round, VS-OPSD (with best-of-$N$ verification gate) yields far more correct rollouts than OPSD and matches or exceeds OPD, evidence that the verification gate, not a stronger teacher, drives the supply of verified self-generated targets.}
    \label{fig:opsd_opd_train_rollout}
\end{figure}

Figure~\ref{fig:opsd_opd_train_rollout} shows the training rollout yields by OPSD, VS-OPSD and OPD respectively. Considering only \verb|OPSD → VS-OPSD|, the gate is a clear win at AECV-2B, mixed at AECV-4B, and neutral-to-negative at AECV-8B. ArchCAD shows a different pattern: the gate helps most at 4B and is roughly neutral at 2B and 8B.

\textbf{Inner-loss ablation: cross-entropy outperforms generalized JSD} 

\begin{figure}   [htbp]
    \centering
    \includegraphics[width=1.0\textwidth]{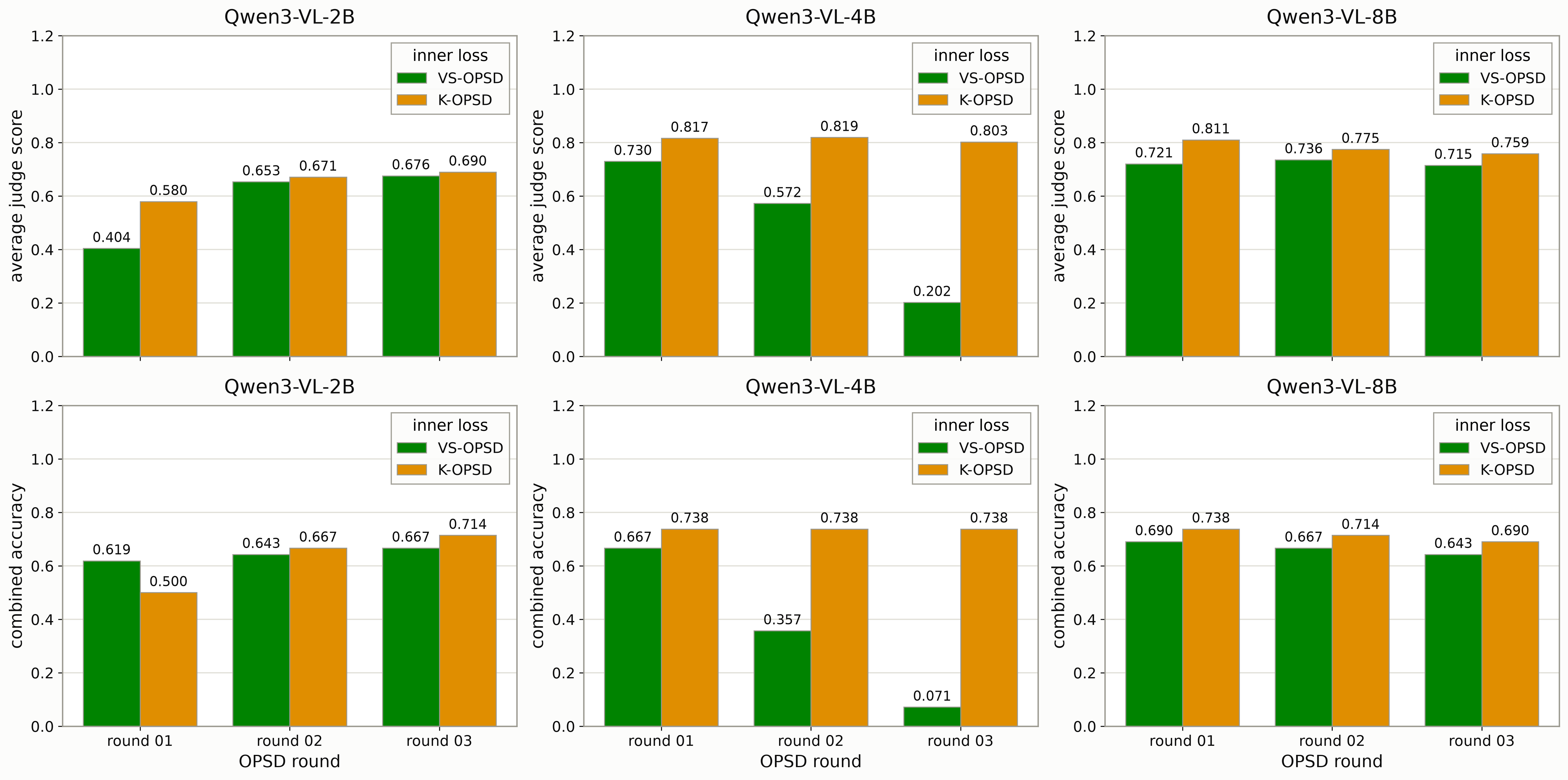}
    \caption{\textbf{Downstream quality by inner-loss loop: VS-OPSD by JSD vs K-OPSD by cross-entropy} across three rounds for Qwen3-VL-2B, 4B, and 8B. Both methods share the identical best-of-$N$ verified teacher construction and differ only in the inner loss. For both average judge score (top) and combined accuracy (bottom), K-OPSD matches or exceeds JSD at nearly every scale and round, and is more stable and more reliable as a distillation objective.}
    \label{fig:inner_loop_accuracy}
\end{figure}

As shown in Figure~\ref{fig:inner_loop_accuracy}, the \verb|VS-OPSD → K-OPSD| pair isolates the inner loss on an identical verified corpus. On AECV-Bench, cross-entropy is never worse than token-wise generalized JSD at $\beta = 0.5$, and is strictly better in five of six results: judge 0.676 → 0.699 (2B), 0.730 → 0.819 (4B), 0.736 → 0.811 (8B), accuracy 0.667 → 0.690 (2B), 0.667 → 0.738 (4B), 0.690 → 0.690 (8B, tied). ArchCAD agrees at 2B (judge 0.497 → 0.589, accuracy 0.293 → 0.341) and 8B (judge 0.599 → 0.693, accuracy 0.317 → 0.366), with the lone exception of ArchCAD-4B, where cross-entropy is marginally lower on both metrics (judge 0.666 → 0.660, accuracy 0.439 → 0.390). This is the empirical basis for K-OPSD and the token-level gradient-saturation analysis. The bounded generalized JSD plateaus on high-confidence-wrong tokens where its corrective logit gradient scales as minimum, while cross-entropy on the verified token retains the gradient.

\vspace{-10pt} 
\section{Conclusion and Future Work}

We studied the adaptation of small vision–language models to AEC drawing understanding and proposed K-OPSD, a verifiable on-policy self-distillation loop for post-training AEC-domain-specific VLMs. K-OPSD builds a teacher from the model’s own verified best-of-$N$ generation sampling, a process-level verifier that certifies the reasoning rather than only the answer, and hint-conditioned rationalization re-verified with the hint removed. When internalizing this verified corpus in the policy, our use of cross-entropy on the selected completions outperforms the bounded token-wise generalized JSD used by on-policy distillation. While our empirical results are limited by the small size of the training and held-out test sets, we observed that K-OPSD fine-tuned Qwen3-VL models attain competitive performance among the compared models on the in-domain AECV-Bench, and the recipe transfers to the out-of-domain ArchCAD set. In future work, we will increase the volume of training data and test on a wider variety of datasets in the domain, and further explore the gradient-saturation mechanism at the per-token level through a $\beta$ sweep, and the rejected best-of-$N$ candidates as a negative learning signal that the present loop discards. We will also conduct quality versus cost analysis to evaluate the sample-efficiency for the OPSD approach.

%\newpage

\section*{References}

{
\small

[1] Lin, J.-R., Cai, Y.-H., Ni, X.-R., Zhou, S.\ \& Pan, P.\ (2026) Qwen-BIM: developing large language model for BIM-based design with domain-specific benchmark and dataset. {\it arXiv preprint} arXiv:2602.20812.

[2] Gao, H., Hartmann, T., Zhong, B., Lia, K.\ \& Luo, H.\ (2025) Domain-specific fine-tuning and prompt-based learning: a comparative study for developing natural language-based BIM information retrieval systems. {\it arXiv preprint} arXiv:2508.05676.

[3] Du, C., Esser, S., Nousias, S.\ \& Borrmann, A.\ (2024) Text2BIM: generating building models using a large language model-based multi-agent framework. {\it arXiv preprint} arXiv:2408.08054.

[4] Wu, Y., Wang, L.\ \& Liu, R.\ (2025) CEQuest: benchmarking large language models for construction estimation. {\it arXiv preprint} arXiv:2508.16081.

[5] Madireddy, S., Gao, L., Din, Z., Kim, K., Senouci, A., Han, Z.\ \& Zhang, Y.\ (2025) Large language model-driven code compliance checking in building information modeling. {\it Electronics} {\bf 14}(11):2146.

[6] Li, Y., Dong, Z.\ \& Shao, Y.\ (2025) DrafterBench: benchmarking large language models for tasks automation in civil engineering. {\it arXiv preprint} arXiv:2507.11527.

[7] Fuchs, S., Witbrock, M., Dimyadi, J.\ \& Amor, R.\ (2024) Using large language models for the interpretation of building regulations. {\it arXiv preprint} arXiv:2407.21060.

[8] Lee, G., Jang, S.\ \& Hyun, S.\ (2024) A generalized LLM-augmented BIM framework: application to a speech-to-BIM system. {\it arXiv preprint} arXiv:2409.18345.

[9] Wan, H., Xu, W., Rosenberg, M., Zhang, J.\ \& Siddika, A.\ (2025) Automatic building code review: a case study. {\it arXiv preprint} arXiv:2510.02634.

[10] Liu, J., Geng, Z., Cao, R., Cheng, L., Bocchini, P.\ \& Cheng, M.\ (2025) A large language model-empowered agent for reliable and robust structural analysis. {\it arXiv preprint} arXiv:2507.02938.

[11] Zhang, X., Iturburu, L., Villamizar, J.N., Liu, X., Salmeron, M., Dyke, S.J.\ \& Ramirez, J.\ (2025) Large language model agent for structural drawing generation using ReAct prompt engineering and retrieval augmented generation. {\it arXiv preprint} arXiv:2507.19771.

[12] Zhang, Y., Wei, S., Huang, Y., Su, Y., Lu, S.\ \& Li, H.\ (2025) SDIGLM: leveraging large language models and multi-modal chain of thought for structural damage identification. {\it arXiv preprint} arXiv:2504.11477.

[13] Deng, Z., Du, C., Nousias, S.\ \& Borrmann, A.\ (2025) BIMgent: towards autonomous building modeling via computer-use agents. {\it arXiv preprint} arXiv:2506.07217.

[14] Youwai, S., Phim, D., Murcia, V.G.\ \& Onas, R.C.\ (2025) Investigating the potential of large language model-based router multi-agent architectures for foundation design automation: a task classification and expert selection study. {\it arXiv preprint} arXiv:2506.13811.

[15] Hostetter, H., Naser, M.Z., Huang, X.\ \& Gales, J.\ (2024) Large language models in fire engineering: an examination of technical questions against domain knowledge. {\it arXiv preprint} arXiv:2403.04795.

[16] Zhang, H., Yan, J., Liu, Q., Yang, J., Su, Y., Li, Z.\ \& Chen, S.\ (2026) BIMVLM: a vision-language model for iterative generation of component BIM models. {\it Expert Systems with Applications}. https://doi.org/10.1016/j.eswa.2026.131765.

[17] Liu, Y., Yang, Z., Li, Y.\ \& Yang, Y.\ (2026) FloorplanVLM: a vision-language model for floorplan vectorization. {\it arXiv preprint} arXiv:2602.06507.

[18] Elshaboury, H., Re Cecconi, F., Scotti, V., Baresi, L.\ \& De Angelis, E.\ (2025) LLM-based processing of design inspection reports as a measure of building design quality. In {\it Proceedings of the 2025 European Conference on Computing in Construction (EC3)}, Porto, Portugal.

[19] Mason \& Hanger\ (2025) Top eight ways artificial intelligence (AI) is transforming building code compliance. [Online]. https://www.masonandhanger.com/news/top-eight-ways-artificial-intelligence-ai-is-transforming-building-code-compliance-part-2.

[20] Mirhosseini, N., Shojaei, D.\ \& Sabri, S.\ (2026) A systematic review of methods for interpreting building code regulations in automated compliance systems. {\it Building Research \& Information}. https://doi.org/10.1080/09613218.2026.2637965.

[21] Zelikman, E., Wu, Y., Mu, J.\ \& Goodman, N.D.\ (2022) STaR: bootstrapping reasoning with reasoning. In {\it Advances in Neural Information Processing Systems 35 (NeurIPS 2022)}. arXiv:2203.14465.

[22] Yuan, Z., Yuan, H., Li, C., Dong, G., Lu, K., Tan, C., Zhou, C.\ \& Zhou, J.\ (2023) Scaling relationship on learning mathematical reasoning with large language models. {\it arXiv preprint} arXiv:2308.01825.

[23] Gulcehre, C., Paine, T.L., Srinivasan, S., Konyushkova, K., Weerts, L., Sharma, A., Siddhant, A., Ahern, A., Wang, M., Gu, C., Macherey, W., Doucet, A., Firat, O.\ \& de Freitas, N.\ (2023) Reinforced self-training (ReST) for language modeling. {\it arXiv preprint} arXiv:2308.08998.

[24] Singh, A., Co-Reyes, J.D., Agarwal, R., Anand, A., Patil, P., Garcia, X., Liu, P.J., Harrison, J., Lee, J., Xu, K., Parisi, A., Kumar, A., Alemi, A., Rizkowsky, A., Nova, A., Adlam, B., Bohnet, B., Elsayed, G., Sedghi, H.\ {\it et al.}\ (2023) Beyond human data: scaling self-training for problem-solving with language models. {\it Transactions on Machine Learning Research (TMLR)}. arXiv:2312.06585.

[25] Agarwal, R., Vieillard, N., Zhou, Y., Stanczyk, P., Ramos, S., Geist, M.\ \& Bachem, O.\ (2024) On-policy distillation of language models: learning from self-generated mistakes. In {\it International Conference on Learning Representations (ICLR 2024)}. arXiv:2306.13649.

[26] Harne, S., Karkar, C., Pandya, Y., Awadallah, A. \& Nambi, A. (2026) Privileged, but biased: how PI-conditioned teachers break self-distillation. arXiv preprint arXiv:2608.04794.

[27] Kaur, S., Ri, N., He, Y., Fowl, L. \& Arora, S. (2026) Rethinking on-policy self-distillation for thinking models. arXiv preprint arXiv:2607.05184.

[28] Kondratenko, A., Birhane, M., Hsain, H.E. \& Maciocci, G. (2026) AECV-Bench: benchmarking multimodal models on architectural and engineering drawings understanding. arXiv preprint arXiv:2601.04819.

[29] Luo, R., Liu, Z., Cheng, T., Wang, J., Wang, T., Wei, X., Wang, H., Li, Y., Chai, F., Cheng, F., Ye, S., Wang, W., Zhang, Y., Qiao, Y., Zhang, H. \& Zhao, X. (2025) ArchCAD-400K: a large-scale CAD drawings dataset and new baseline for panoptic symbol spotting. arXiv preprint arXiv:2503.22346.

}

\appendix

\section{Limitations}
\label{app:limitations}

This work is a preliminary study offering feasibility-level evidence that verifier-guided self-distillation is effective for AEC drawing understanding. We intend to extend it into a more comprehensive study with larger datasets, broader model families, and the verifier, cost, and mechanism analyses.

\textbf{Sample size and statistical power} The pipeline is trained on a small corpus (100 SFT samples and 150 self-distillation samples) and evaluated on held-out sets of only 42 (AECV-Bench) and 40 (ArchCAD) items. It is encouraging that the recipe delivers competitive results from the compact corpus, that is evidence of strong data efficiency. We read the ordering among OPD, OPSD, VS-OPSD, and K-OPSD and the cross-entropy-versus-JSD comparison as promising trends, and the next steps are to confirm them with larger held-out sets, multiple seeds, confidence intervals and paired significance tests, which the pipeline readily accommodates.

\textbf{Empirical study of the gradient-saturation mechanism} Our explanation for why cross-entropy improves on the bounded token-wise generalized JSD is grounded in a clean token-level derivation, and the ablation results are consistent with it. A compelling way to further corroborate the mechanism is to probe per-token gradient norms and loss in the high-confidence-wrong regime and to sweep $\beta$ for the generalized JSD. The present experiments fix $\beta = 0.5$, turning a well-motivated theoretical account into a directly measured one.  

\textbf{Quality–cost analysis} Because a small domain-specialized model matches much larger baselines, the efficiency story is a particular strength of this recipe, and quantifying it is a valuable extension rather than a gap in the core method. Reporting training cost (GPU-hours per round, rollouts consumed, rounds used), inference cost relative to the frontier baselines, and a quality-versus-rollout-budget curve will make the "sample-efficient" and "practical paradigm" framing concrete and, we expect, favorable.

\textbf{Verifier and human-in-the-loop characterization} The process-level verifier is the engine of the method, and detailing its reliability against subject-matter-expert (SME) judgment (precision, recall, false-positive rate), its coverage across question types, and worked accept/reject examples will further strengthen an already central contribution. Likewise, measuring the scope and override behavior of the human-in-the-loop review will let us quantify precisely how lightweight the human oversight is, reinforcing the autonomy of the loop.

\textbf{Broader evaluation and baselines} The results are reported with two complementary metrics: a deterministic Combined Accuracy and an LLM-judge score from two independent with different-family judges. The future versions can further shore up the evaluation with judge–human and inter-judge agreement, weight-sensitivity analysis, and a compute- or tuning-matched comparison. 

\begin{figure} %[htbp]
    \centering
    \includegraphics[width=1.0\textwidth]{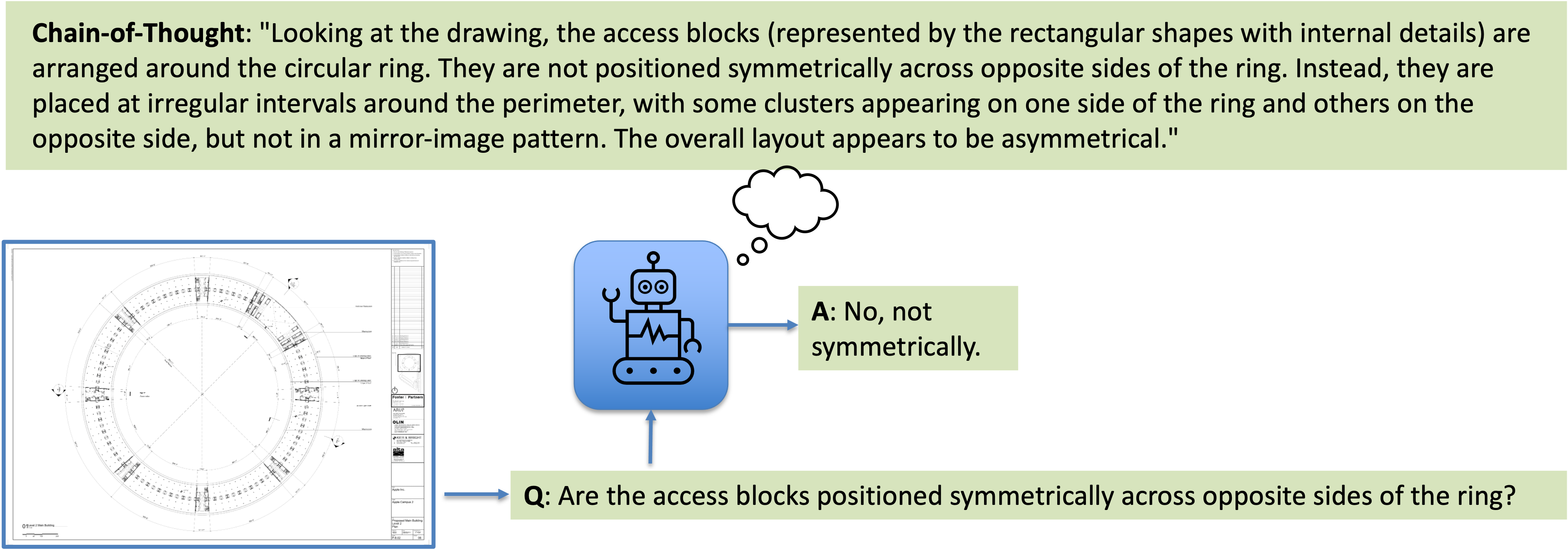}
    \caption{\textbf{Example of VQA on AEC drawing}: given an engineering drawing (left) and a query (bottom), the model localizes the relevant elements, examines the spatial distribution, and performs visual reasoning (top) that enables the model to provide an answer (right) for complex, real-world diagrams.}
    \label{fig:figure_1}
\end{figure}

\section{VQA task for AEC drawing}

We define AEC drawing understanding as a visual question answering (VQA) task. The MMLM/VLM is given a rendered drawing, such as a floor plan, section, or construction detail, together with a natural-language question. The model must ground its answer in the visual content of the drawing. More than just reading text alone, the model localizes the relevant elements (for example, symbols, dimension lines, annotations, and geometry), interprets the drawing's symbolic and spatial conventions, and reasons over them to produce an answer. The output includes an explicit reasoning trace in its Chain-of-Thought, followed by the extracted answer. This makes the model handle heterogeneous query types, such as dimensional facts, symbol and entity identification, BIM-schema values, and building-code judgments, from the same visual input. 

This is the setting in which K-OPSD's verifiable supervision applies because each answer can be checked against the drawing's ground truth. The example and prompt in Figure~\ref{fig:figure_1} illustrate how the drawing and question are presented to the model and how it structures its visual reasoning and response.

\section{K-OPSD training illustration}

\subsection{Algorithm}

% Requires in the preamble:  \usepackage{algorithm}  and  \usepackage{algorithmic}
% (classic algorithmic; do NOT also load algpseudocode or algorithm2e — they clash)
\begin{algorithm} 
\caption{K-OPSD Training Pipeline}
\label{alg:vsopsd}
\begin{algorithmic}[1]
\REQUIRE dataset $\mathcal{D} = \{(x, y)\}$, frozen SFT-merged base $\theta_0$, rounds $R$, samples per prompt $N$, process-level verifier $V$, inner loss 
\ENSURE final adapter $\phi^\star$ (loaded on top of $\theta_0$)
\STATE $\phi_0 \gets \emptyset$ \COMMENT{round 1 student is the bare SFT base}
\FOR{$r = 1$ to $R$}
    \STATE $\pi \gets \theta_0 \oplus \phi_{r-1}$ \COMMENT{current on-policy student, $\oplus$ = base with adapter}
    \STATE $\mathcal{T} \gets \emptyset$ \COMMENT{verified teacher targets for this round}
    \FORALL{$(x, y) \in \mathcal{D}$}
        \STATE $\mathcal{C} \gets \{\, c_i = (r_i, \hat{y}_i) \sim \pi(\cdot \mid x) : i = 1, \dots, N \,\}$ \COMMENT{(1) on-policy rollout}
        \STATE $\mathcal{C}^{+} \gets \{\, c \in \mathcal{C} : V(x, c, y) = 1 \,\}$ \COMMENT{(2) verify \& select}
        \IF{$\mathcal{C}^{+} \neq \emptyset$}
            \STATE $c^\star \gets \mathrm{BestOf}(\mathcal{C}^{+})$
            \STATE $\mathcal{T} \gets \mathcal{T} \cup \{(x, c^\star)\}$ \COMMENT{keep a verified completion}
        \ELSE
            \STATE $\tilde{c} \sim \pi\big(\cdot \mid x, \mathrm{Hint}(y)\big)$ \COMMENT{(3) hinted rescue, hint visible at sample time only}
            \IF{$V(x, \tilde{c}, y) = 1$ \textbf{and} $\lnot\,\mathrm{Leaked}(\tilde{c})$}
                \STATE $\mathcal{T} \gets \mathcal{T} \cup \{(x, \mathrm{StripHint}(\tilde{c}))\}$ \COMMENT{(4) strip hint from target}
            \ENDIF
        \ENDIF
    \ENDFOR
    \STATE $\phi_r \gets \arg\min_{\phi} \; \mathcal{L}\big(\theta_0 \oplus \phi; \, \mathcal{T}\big)$ \COMMENT{(5) fresh adapter on frozen base, eval + early stop}
    \STATE $s_r \gets \mathrm{Eval}(\theta_0 \oplus \phi_r,\, \mathcal{D}_{\mathrm{val}})$ \COMMENT{(6) held-out eval: post vs same-round pre}
\ENDFOR
\STATE $\phi^\star \gets \phi_{\,\arg\max_r s_r}$ \COMMENT{promote the best round}
\RETURN $\phi^\star$
\end{algorithmic}
\end{algorithm}

We describe the concrete procedure that realizes the training pipeline in Algorithm 1, which one round $r$ maps the previous round's adapter $\phi_{r-1}$ to a new adapter $\phi_r$ through six steps. The adapter $\phi_r$ becomes the student policy for round $r+1$, making the rollouts on-policy and cumulative. Each round corrects errors on the model's own updated distribution and raises the ceiling for the next round. After $R$ rounds the best adapter is promoted by held-out score $\phi^\star \;=\; \phi_{\,\arg\max_r \, s_r}$, and loaded on top of $\theta_0$ as the final model. Within this loop the verifier suite is the selection of Step 2 and the re-verification of Step 4, which jointly govern which samples and which rescued traces become teacher targets. 

\subsection{Hint-conditioned self-teacher prompt versus external-teacher prompt}

The following example shows the three prompts used in K-OPSD training on the same drawing and question: the unaided rollout prompt used for best-of-N generation, the hint-conditioned prompt used only on the rescue path, and the external-teacher prompt used in the OPD baseline.

\fcolorbox{black} 
{yellow!20}{% % Border color, Background color
\parbox{13cm}{\small
\medskip
 
\textbf{Question}: How many labeled access blocks are distributed around the building ring?  \\ \\
\textbf{Answer}: Nine (9) \\ 

\textbf{Rollout Prompt}: \\
You are analyzing a high-resolution crop taken from a construction diagram. \\ \\
The crop was already zoomed into the region relevant to the user's question.\\
 - Read all visible text, tags, labels, dimensions, and symbols in the crop.\\
 - Answer the user's question using only what you can see in the crop.\\
 - If the crop does not contain enough information, say so in a few words.\\
 - Analyze the image carefully. Let's think step by step to find the answer, detailing your observations and logical deductions before concluding.\\ \\
Response format:\\
 - Put your reasoning inside \texttt{<think></think>} tags.\\
 - Put ONLY the final answer inside \texttt{<answer></answer>} tags.\\
 - Keep your reasoning SHORT and focused - a few brief sentences at most. Do not restate the question, list every symbol, second-guess yourself, or repeat points. Over-long reasoning can run out of space before the answer.\\
 - IMPORTANT: You MUST close \texttt{</think>} and then give the \texttt{<answer></answer>} block. The answer is required. As soon as you know the answer, stop reasoning, close \texttt{</think>}, and write the \texttt{<answer>}. Never let the reasoning consume all available space.\\
 - Answer directly and as briefly as possible, a tag, number, name, or single short phrase where possible. Give the answer straight away with no preamble, no restating of the question, no explanation, no justification, and no closing remarks.\\

\textbf{Hint-Conditioned Self-Teacher Prompt}: \\
Here is a construction drawing image relevant to the question: [<IMAGE>] How many labeled access blocks are distributed around the building ring? \\ \\
REFERENCE - background for your reasoning only, not part of the question   \\
Correct answer: Nine (9)   \\
Evidence: From the drawing: Nine rectangular core blocks appear around the circular ring.   \\
Task: find this in the image yourself, then write the reasoning that leads to it.    \\
 - In \texttt{<think>}, cite ONLY what you can actually see: labels, tags, dimensions, room names, symbols, and their positions.    \\
 - Write as if you had worked it out unaided. Do NOT mention this reference, and do NOT use words like hint, given, provided, verified, or reference.   \\
 - Put the final answer in \texttt{<answer>}, as brief as the correct answer above.   \\

\textbf{External-Teacher Prompt}: \\
Here is a construction drawing image relevant to the question: [<IMAGE >] How many labeled access blocks are distributed around the building ring?   \\
You are an expert construction-drawing analyst writing a training demonstration. The verified correct answer to the question is: Nine (9)  \\
Reference evidence (for your reasoning only): From the drawing: Nine rectangular core blocks appear around the circular ring.  \\
Write the reasoning that leads to that answer AS IF you worked it out unaided from the image:   \\
- In \texttt{<think>}, cite ONLY what is actually visible: labels, tags, dimensions, room names, symbols, counts, and their positions. For a count, enumerate the instances. For a code, quote it verbatim. For a direction, name a landmark and a direction word.  \\   
- Do NOT mention that an answer was provided; do NOT use the words hint, given, provided, verified, reference, or 'the answer is provided'.  \\
- Put ONLY the final answer in \texttt{<answer>}, as brief as the correct answer above.   

\medskip
}
}

All three prompts issue the same drawing-and-question VQA task under the same \texttt{<think>/<answer>} output contract. The OPD/OPSD axis is the provenance of the candidate generations: OPD draws them from an external model, OPSD from the student's own policy. Privileged information is an orthogonal dimension. K-OPSD generates with the unaided rollout prompt by default and uses the hint-conditioned prompt only on the rescue path, where the hint is stripped and the trace independently re-verified under the unconditioned prompt. OPD with the external-teacher prompt is answer-conditioned and, coming from a different model, off-policy. Answer-conditioned generation raises coverage and answer accuracy but tends to produce "right answer, wrong reasoning" traces that bias the student toward the reference [26, 27], which is why an accepted rescue trace must clear the process-level verifier and the leakage guard rather than an answer check alone. Consistent with this, the ablation in Section 4.3 shows the verification-gated self-teacher matching or exceeding the external-teacher OPD on judge score at every scale, with OPD retaining only a small combined-accuracy edge in a few cases.

\subsection{Train rollout: success rollout versus hinted rollout across rounds}

\begin{figure}   [htbp]
    \centering
    \includegraphics[width=1.0\textwidth]{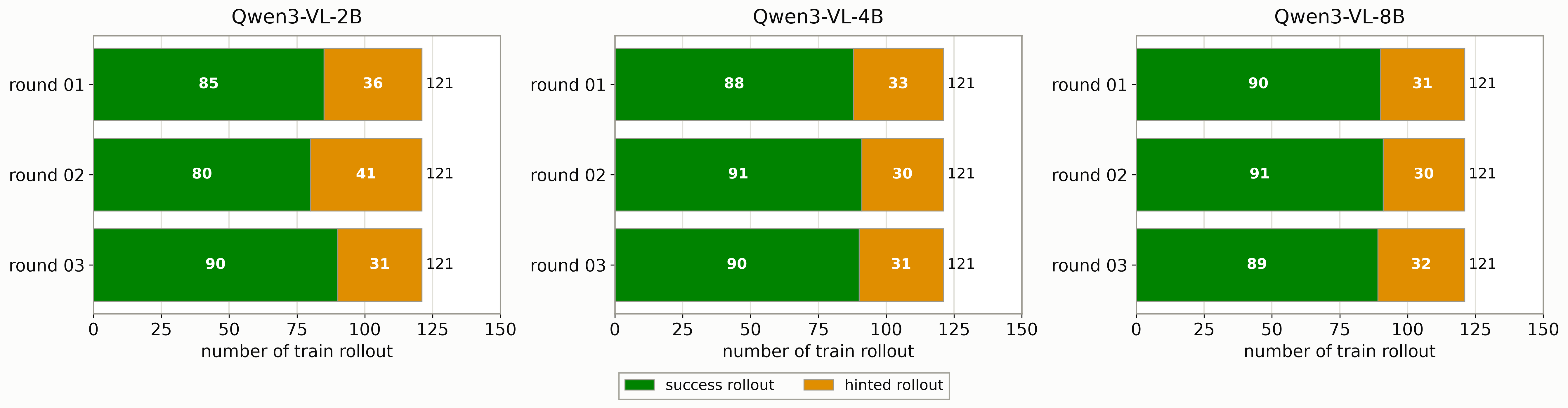}
    \caption{\textbf{K-OPSD train rollout}: composition of the 121-prompt training pool per K-OPSD round for Qwen3-VL-2B, 4B, and 8B. Each bar splits the pool into prompts solved unaided by the best-of-$N$ rollout (success, green) vs those failed and sent to hinted rescue (orange)}
    \label{fig:k-opsd_train_rollout}
\end{figure}

Unaided success dominates the training pool and is largely flat across rounds and scales. For every (model, round) the stacked bar sums to exactly 121, that confirms the pipeline's invariant that each prompt the best-of-$N$ rollout fails is sent to hinted rescue, so success + hinted partitions the full pool with no leakage. Details are presented in Figure~\ref{fig:k-opsd_train_rollout}.

Across all (model, round) the split sits in a narrow band: roughly 80–91 solved unaided against 30–41 sent to rescue. Scale buys only a small, mostly stability-related gain. The 4B and 8B runs hold steady at 88–91 unaided every round, while the smallest 2B model is both slightly lower and noisier (85 → 80 → 90, dipping in round 02). Notably, success does not climb monotonically with rounds at any scale, it oscillates within a few prompts rather than trending up, indicating the unaided solve rate has effectively saturated near a per-prompt-pool ceiling and that the round-over-round gains in the model come from refining the rescued prompts rather than from steadily converting more of the pool to unaided success.

\section{The method ladder in ablation study}

\begin{table} [htbp]
    \centering
\caption{\textbf{The method ladder used in the controlled ablation}, characterized along three axes: the teacher vs the model's own on-policy self-teacher, with/without process-level verification gate with best-of-$N$, and the inner loss that internalizes the weights update using generalized JSD versus cross-entropy on the verified completion.}
\label{tab:ladder}
    \begin{tabular}{cccc}\toprule
         Method&  Teacher model&  Verification Gate&  Inner-loop Loss\\\midrule
         OPD&  External (Claude Fable 5)&  None&  generalized JSD\\
         OPSD&  On-policy self-teacher&  None&  generalized JSD\\
         VS-OPSD&  On-policy self-teacher&  process-level verifier&  generalized JSD\\
 \textbf{K-OPSD (Ours)}& On-policy self-teacher& process-level verifier& \makecell{Cross-entropy}\\ \bottomrule
    \end{tabular}
\end{table}

%\begin{figure}   %[htbp]
%    \centering
%    \includegraphics[width=1.0\textwidth]{Figures/opsd_jsd_loss.png}
%    \caption{JSD loss OPSD vs VS-OPSD}
%    \label{fig:opsd_jsd_loss}
%\end{figure}

\newpage
\section{LLM-judge Prompt}
\label{app:prompts}

%\begin{verbatim}
%\noindent
\fcolorbox{black}
{yellow!20}{% % Border color, Background color 
\parbox{13cm}{\small
\medskip
You are an expert evaluator for questions about construction/engineering drawings. \\

You will be given: \\
- A question \\
- The ground truth answer \texttt{(gt\_answer)} \\
- A model's answer to evaluate \texttt{(model\_answer)} \\
- The model's reasoning / chain-of-thought \texttt{(model\_cot / model\_reasoning)}, which may be empty.  \\ 

Rate the model on three factors, each on a 0.0-1.0 scale, then combine them into a single weighted accuracy score:  \\
  1. answer\_accuracy \texttt{(weight 0.75)}: the level of alignment between model\_answer and gt\_answer.  \\
    - 1.0 = states the same fact even if phrased, formatted, or contextualized differently \texttt{(e.g. "2" vs "Two (2)", "fridge" vs "refrigerator", units dropped)};  \\
    - partial credit \texttt{(e.g. 0.5)} when it captures some but not all of the fact;  \\
    - 0.0 = a different value/name/direction or unrelated answer \texttt{(e.g. "R-36" vs "R-38", "top-right" vs "bottom-right")}. \\
  2. format \texttt{(weight 0.1)}:  \\
    - 1.0 if BOTH model\_answer and model\_cot are available (present and non-empty);  \\
    - 0.0 if either is missing.  \\
  3. reasoning \texttt{(weight 0.15)}:  \\
    - 1.0 if model\_answer is based on model\_reasoning, i.e. the answer logically follows from and is grounded in the reasoning;  \\
    - partial credit if only loosely supported;  \\
    - 0.0 if the reasoning is absent, contradicts the answer, or is irrelevant. \\
  
Compute the overall score as the weighted sum: \\
  \texttt{score = 0.75 * answer\_accuracy + 0.1 * format + 0.15 * reasoning} \\
  
Respond with ONLY a JSON object in this exact format: \\
  \{"answer\_accuracy": \texttt{<0.0-1.0>}, "format": \texttt{<0.0-1.0>}, "reasoning": \texttt{<0.0-1.0>},  \\
   "score": \texttt{<weighted sum, 0.0-1.0>}, "explanation": "\texttt{<one sentence>}"\} \\
\medskip
}
}  
%\end{verbatim}
\\

%%%%%%%%%%%%%%%%%%%%%%%%%%%%%%%%%%%%%%%%%%%%%%%%%%%%%%%%%%%%

%\newpage
%\input{checklist.tex}

\end{document}